\documentclass{article} 
\usepackage{iclr2017_conference,times}
\usepackage{hyperref}
\usepackage{url}
\usepackage{booktabs}
\usepackage{multirow}
\usepackage{graphicx}
\usepackage{amsfonts}
\usepackage{rotating}
\usepackage{lettrine}

\def \ie {\emph{i.e.}}
\def \eg {\emph{e.g.}}

\def \mys {\hspace{7pt}}
\def \mysb {\hspace{9pt}}

\title{FastText.zip:\newline
  Compressing text classification models}

\author{Armand Joulin, Edouard Grave, Piotr Bojanowski, Matthijs Douze, Herv\'e J\'egou \& Tomas Mikolov \\
Facebook AI Research \\
\texttt{\{ajoulin,egrave,bojanowski,matthijs,rvj,tmikolov\}@fb.com} \\
}

\begin{document}

\maketitle

\begin{abstract}
We consider the problem of producing compact architectures for text
classification, such that the full model fits in a limited amount of memory.
After considering different solutions inspired by the hashing literature, we
propose a method built upon product quantization to store word
embeddings. While the original technique leads to a loss in
accuracy, we adapt this method to circumvent quantization artefacts. 
%
Our experiments carried out on several benchmarks show that our approach
typically requires two orders of magnitude less memory than \texttt{fastText} while
being only slightly inferior with respect to accuracy. 
As a result, it outperforms the state of the art by a good margin in terms of the compromise
between memory usage and accuracy.   

\end{abstract}

\section{Introduction}

Text classification is an important problem in Natural Language Processing (NLP). 
Real world use-cases include spam filtering or e-mail categorization. 
It is a core component in more complex systems such as search and ranking. Recently, deep
learning techniques based on neural networks have achieved state of the art
results in various NLP applications. 
One of the main successes of deep learning is due to the effectiveness of
recurrent networks for language modeling and their application to speech
recognition and machine translation~\citep{mikolov2012}. However, in other
cases including several text classification problems, it has been shown that
deep networks do not convincingly beat the prior state of the art
techniques~\citep{W12,J16}.

In spite of being (typically) orders of magnitude slower to train than traditional
techniques based on n-grams, neural networks are often regarded as a 
promising alternative due to compact model sizes, in particular for character based
models. This is important for applications that need to run on systems
with limited memory such as smartphones. 

This paper specifically addresses the compromise between classification accuracy and the model size. 
We extend our previous work implemented in the \texttt{fastText} library\footnote{https://github.com/facebookresearch/fastText}. It is based on n-gram
features, dimensionality reduction, and a fast approximation of the softmax classifier~\citep{J16}. 
We show that a few key ingredients, namely feature pruning, quantization, hashing, and re-training, allow us to produce text classification models with tiny size, often less than 100kB when trained on several popular datasets, without noticeably sacrificing accuracy or speed. 

We plan to publish the code and scripts required to
reproduce our results as an extension of the \texttt{fastText} library, thereby providing strong reproducible
baselines for text classifiers that optimize the compromise between the model size and accuracy. 
We hope that this will help the engineering community to improve existing applications by using more efficient models.

This paper is organized as follows. Section~\ref{sec:related} introduces related work, Section~\ref{sec:approach} describes our text classification model and explains how we drastically reduce the model size. Section~\ref{sec:experiments} shows the effectiveness of our approach in experiments on multiple text classification benchmarks. 

\section{Related work}
\label{sec:related}

\paragraph{Models for text classification.}
Text classification is a problem that has its roots in many applications such
as web search, information retrieval and document
classification~\citep{deerwester1990indexing,pang2008opinion}.  Linear
classifiers often obtain state-of-the-art performance while being
scalable~\citep{A14,joachims1998text,J16,mccallum1998comparison}.  They are
particularly interesting when associated with the right features~\citep{W12}.
They usually require storing embeddings for words and n-grams, which makes
them memory inefficient.

\paragraph{Compression of language models.}
Our work is related to compression of statistical language models. Classical approaches
include feature pruning based on entropy~\citep{stolcke2000entropy} and quantization.
Pruning aims to keep only the most important n-grams in the model, leaving out those
with probability lower than a specified threshold. Further, the individual n-grams can be compressed
by quantizing the probability value, and by storing the n-gram itself more efficiently than as a sequence
of characters. Various strategies have been developed, for example using tree structures or hash functions,
and are discussed in~\citep{talbot2008randomized}.

\paragraph{Compression for similarity estimation and search.} 
There is a large body of literature on how to compress a set of vectors into compact codes, such that the comparison of two codes approximates a target similarity in the original space. The typical use-case of these methods considers an indexed dataset of compressed vectors, and a query for which we want to find the nearest neighbors in the indexed set.  
One of the most popular is Locality-sensitive hashing (LSH) by~\citet{C02}, which is a binarization technique based on random projections that approximates the cosine similarity between two vectors through a monotonous function of the Hamming distance between the two corresponding binary codes. In our paper, LSH refers to this binarization strategy\footnote{In the literature, LSH refers to multiple distinct strategies related to the Johnson-Lindenstrauss lemma. For instance, LSH sometimes refers to a partitioning technique with random projections allowing for sublinear search \emph{via} cell probes, see for instance the E$^2$LSH variant of~\citet{DIIM04}.}.
Many subsequent works have improved this initial binarization technique, such as spectal hashing~\citep{WTF09}, or Iterative Quantization (ITQ)~\citep{GL11}, which learns a rotation matrix minimizing the
quantization loss of the binarization. We refer the reader to two recent surveys by~\citet{WSSJ14} and \citet{WLKC15} for an overview of the binary hashing literature. 

Beyond these binarization strategies,  more general quantization techniques derived from~\citet{JDS11} offer better trade-offs between memory and the approximation of a distance estimator. 
The Product Quantization (PQ) method approximates the distances by calculating, in the compressed domain, the distance between their quantized approximations. 
This method is statistically guaranteed to preserve the Euclidean distance between the vectors within an error bound directly related to the quantization error.
The original PQ has been concurrently improved by~\citet{GHKS13} and~\citet{NF13}, who learn an orthogonal transform minimizing the overall quantization loss. In our paper, we will consider the Optimized Product Quantization (OPQ) variant~\citep{GHKS13}. 

\paragraph{Softmax approximation} 
The aforementioned works approximate either the Euclidean distance or the cosine similarity (both being equivalent in the case of unit-norm vectors). However, in the context of \texttt{fastText}, we are specifically interested in approximating the maximum inner product involved in a softmax layer. Several approaches derived from LSH have been recently proposed to achieve this goal, such as Asymmetric LSH by \citet{SL14}, subsequently discussed by~\citet{NS15}. 
In our work, since we are not constrained to purely binary codes, we resort a more traditional encoding by employing a magnitude/direction parametrization of our vectors. Therefore we only need to encode/compress an unitary d-dimensional vector, which fits the aforementioned LSH and PQ methods well. 

\paragraph{Neural network compression models.} Recently, several research efforts have been conducted to compress the parameters of architectures involved in computer vision, namely for state-of-the-art Convolutional Neural Networks (CNNs)~\citep{HMD16,LCMB15}. Some use vector quantization~\citep{GLYB14} while others binarize the network~\citep{CHSEB16}. \citet{DSDRF13} show that such classification models are easily compressed because they are over-parametrized, which concurs with early observations by~\citet{LDS90}.  

Some of these works both aim at reducing the model size and the speed. In our case, since the \texttt{fastText} classifier on which our proposal is built upon is already very efficient, we are primilarly interested in reducing the size of the model while keeping a comparable classification efficiency. 

\section{Proposed approach}
\label{sec:approach}

\subsection{Text classification}

In the context of text classification, linear classifiers~\citep{J16} remain competitive with more sophisticated, deeper models, and are much faster to train. On top of standard tricks commonly used in linear text classification~\citep{A14,W12,W09}, \citet{J16} use a low rank constraint to reduce the 
computation burden while sharing information between different classes. 
This is especially useful in the case of a large output space, where rare classes may have only a few training examples. 
In this paper, we focus on a similar model, that is, which minimizes the softmax loss $\ell$ over $N$ documents:
\begin{eqnarray}\label{eq:fasttext}
\sum_{n=1}^N \ell(y_n, BAx_n),
\end{eqnarray}
where $x_n$ is a bag of one-hot vectors and $y_n$ the label of the $n$-th document.
In the case of a large vocabulary and a large output space, the matrices $A$ and $B$ are big and can require  gigabytes of memory. Below, we describe how we reduce this memory usage. 

\subsection{Bottom-up product quantization}
\label{sec:retrain}

\paragraph{Product quantization} is a popular method for compressed-domain approximate nearest neighbor search~\citep{JDS11}. As a compression technique, it approximates a real-valued vector by finding the closest vector in a pre-defined structured set of centroids, referred to as a codebook. This codebook is not enumerated, since it is extremely large. Instead it is implicitly defined by its structure:   
a $d$-dimensional vector $x \in {\mathbb R}^d$ is approximated as
\begin{equation}
\hat{x} = \sum_{i=1}^k q_i(x), 
\end{equation}
where the different subquantizers $q_i:x \mapsto q_i(x)$ are complementary in
the sense that their respective centroids lie in distinct orthogonal subspaces,
\ie, $\forall i \ne j,\ \forall x,y,\ \langle q_i(x) | q_j(y) \rangle = 0$.
In the original PQ, the subspaces are aligned with the natural axis, while
OPQ learns a rotation, which amounts to alleviating this constraint and to
not depend on the original coordinate system. Another way to see this is to
consider that PQ splits a given vector $x$ into $k$ subvectors $x^i$,
$i=1\dots k$, each of dimension $d/k$:  $x=[x^1 \dots x^i \dots x^k]$, and
quantizes each sub-vector using a distinct k-means quantizer. Each
subvector $x^i$ is thus mapped to the closest centroid amongst $2^b$
centroids, where $b$ is the number of bits required to store the
quantization index of the subquantizer, typically $b=8$. The reconstructed
vector  can take $2^{kb}$ distinct reproduction values, and is stored in
$kb$ bits.  

PQ estimates the inner product in the compressed domain as
\begin{equation}
x^\top y  \approx \hat{x}^\top y = \sum_{i=1}^k q_i(x^i)^\top y^i.
\end{equation}

This is a straightforward extension of the square L2 distance estimation of~\citet{JDS11}.
In practice, the vector estimate $\hat{x}$ is trivially reconstructed from the codes, \ie, from the quantization indexes, by concatenating these centroids. 
 
The two parameters involved in PQ, namely the number of subquantizers $k$ and the number of bits~$b$ per quantization index, are typically set to $k \in [2,d/2]$, and $b=8$ to ensure byte-alignment. 

\paragraph{\it Discussion.} 
PQ offers several interesting properties in our context of text classification. Firstly, the training is very fast because the subquantizers have a small number of centroids, \ie, 256 centroids for $b=8$. Secondly, at test time it allows the reconstruction of the vectors with almost no computational and memory overhead. 
Thirdly, it has been successfully applied in computer vision, offering much better performance than binary codes, which makes it a natural candidate to compress relatively shallow models. As observed by \cite{SP11}, using PQ just before the last layer incurs a very limited loss in accuracy when combined with a support vector machine.


In the context of text classification, the norms of the vectors are widely spread, typically with a ratio of 1000 between the max and the min. 
Therefore kmeans performs poorly because it optimizes an absolute error objective, so it maps all low-norm vectors to 0. 
A simple solution is to separate the norm and the angle of the vectors and to quantize them separately. 
This allows a quantization with no loss of performance, yet requires an extra $b$ bits per vector.

\paragraph{Bottom-up strategy: re-training.}
The first works aiming at compressing CNN models like the one proposed
by~\citep{GLYB14} used the reconstruction from off-the-shelf PQ, \ie,
without any re-training. However,  as observed in \citet{S16}, when using
quantization methods like PQ, it is better to re-train the layers
occurring after the quantization, so that the network can re-adjust itself to
the quantization. There is a strong argument arguing for this re-training
strategy: the square magnitude of vectors is reduced, on average, by the
average quantization error for any quantizer satisfying the Lloyd conditions;
see~\citet{JDS11} for details.  

This suggests a bottom-up learning strategy where we first quantize the input
matrix, then retrain and quantize the
output matrix (the input matrix being frozen). Experiments in
section~\ref{sec:experiments} show that it is worth adopting this strategy.  

\paragraph{Memory savings with PQ.}  
In practice, the bottom-up PQ strategy offers a compression factor of 10
without any noticeable loss of performance. Without re-training, we notice a
drop in accuracy between $0.1\%$ and $0.5\%$, depending on the dataset and
setting; see Section~\ref{sec:experiments} and the appendix.

\subsection{Further text specific tricks}
\label{sec:tricks}

The memory usage strongly depends on the size of the vocabulary, which can be large in many text classification tasks. 
While it is clear that a large part of the vocabulary is useless or redundant, directly reducing 
the vocabulary to the most frequent words is not satisfactory: most of the frequent words, like ``the'' or ``is'' are not discriminative, in contrast to some rare words, \eg, in the context of tag prediction.
In this section, we discuss a few heuristics to reduce the space taken by the dictionary. 
They lead to major memory reduction, in extreme cases by a factor $100$. 
We experimentally show that this drastic reduction is complementary with the PQ compression method, meaning that the  combination of both strategies reduces the model size by a factor up to $\times 1000$ for some datasets. 

\paragraph{Pruning the vocabulary.}
Discovering which word or n-gram must be kept to preserve the overall
performance is a feature selection problem.  While many approaches have been
proposed to select groups of variables during training~\citep{BJ12,M08}, we are
interested in selecting a fixed subset of $K$ words and ngrams from a
pre-trained model. This can be achieved by selecting the $K$ embeddings that
preserve as much of the model as possible, which can be reduced to selecting
the $K$ words and ngrams associated with the highest norms.

While this approach offers major memory savings, it has one drawback occurring in some particular cases: some
documents may not contained any of the $K$ best features, leading to a
significant drop in performance. It is thus important to keep the $K$ best
features under the condition that they cover the whole training set.  More
formally, the problem is to find a subset ${\mathcal S}$ in the feature set ${\mathcal V}$ that maximizes 
the sum of their norms $w_s$ under the constraint that all the documents in the
training set ${\mathcal D}$ are covered:
\begin{eqnarray*}
\max_{{\mathcal S}\subseteq \mathcal V} \sum_{s\in \mathcal S} w_s &\textrm{~~~s.t.~~~} & |{\mathcal S}|\le K,~~~P1_{\mathcal S}\ge 1_{\mathcal D},\\
\end{eqnarray*}
where $P$ is a matrix such that $P_{ds}=1$ if the $s$-th feature is in the
$d$-th document, and $0$ otherwise. 
This problem is directly related to set covering
problems that are NP-hard~\citep{feige1998threshold}. 
Standard greedy approaches require the storing of an inverted index or to do
multiple passes over the dataset, which is prohibitive on very large
dataset~\citep{chierichetti2010max}. 
This problem can be cast as an instance of online submodular
maximization with a rank constraint~\citep{badanidiyuru2014streaming,bateni2010submodular}.  In our
case, we use a simple online parallelizable greedy approach: For each
document, we verify if it is already covered by a retained feature and, 
if not, we add the feature with the highest norm to our set of retained
features. If the number of features is below $k$, we add the features with the
highest norm that have not yet been picked. 

\paragraph{Hashing trick \& Bloom filter.}
On small models, the dictionary can take a significant portion of the memory.
Instead of saving it, we extend the hashing trick used in~\cite{J16} to both
words and n-grams.  This strategy is also used in Vowpal Wabbit~\citep{A14} in the context of online training. 
This allows us to save around 1-2Mb with almost no overhead at test time (just the cost of computing the hashing function).

Pruning the vocabulary while using the hashing trick requires keeping a list of
the indices of the $K$ remaining buckets.  At test time, a binary search over
the list of indices is required. It has a complexity of $O(\log(K))$ and a
memory overhead of a few hundreds of kilobytes.  Using Bloom filters instead reduces
the complexity ${\mathcal O}(1)$ at test time and saves a few hundred kilobytes. However,
in practice, it degrades performance.

\section{Experiments}
\label{sec:experiments}

This section evaluates the quality of our model compression pipeline and compare it to other
compression methods on different text classification problems, and to other compact text classifiers.

\paragraph{Evaluation protocol and datasets.}~Our experimental pipeline is as follows: 
we train a model using
\texttt{fastText} with
the default setting unless specified otherwise. That is $2$M buckets, a
learning rate of $0.1$ and $10$ training epochs. The dimensionality $d$ of the
embeddings is set to powers of $2$ to avoid border effects that could make the
interpretation of the results more difficult.  As baselines, we use
Locality-Sensitive Hashing (LSH)~\citep{C02}, PQ~\citep{JDS11} and
OPQ~\citep{GHKS13} (the non-parametric variant).  Note that we
use an improved version of LSH where random orthogonal matrices are used
instead of random matrix projection~\cite{JDS08}.  In a first series of experiments, we use
the $8$ datasets and evaluation protocol of~\citet{Z15}.  These datasets
contain few million documents and have at most $10$ classes. We also explore
the limit of quantization on a dataset with an extremely large output space,
    that is a tag dataset extracted from the YFCC100M collection~\citep{T15}\footnote{Data available at
  https://research.facebook.com/research/fasttext/}, referred to as FlickrTag in the rest of this paper.

\subsection{Small datasets}

\begin{figure}[t]
\centering
\includegraphics[width=\textwidth]{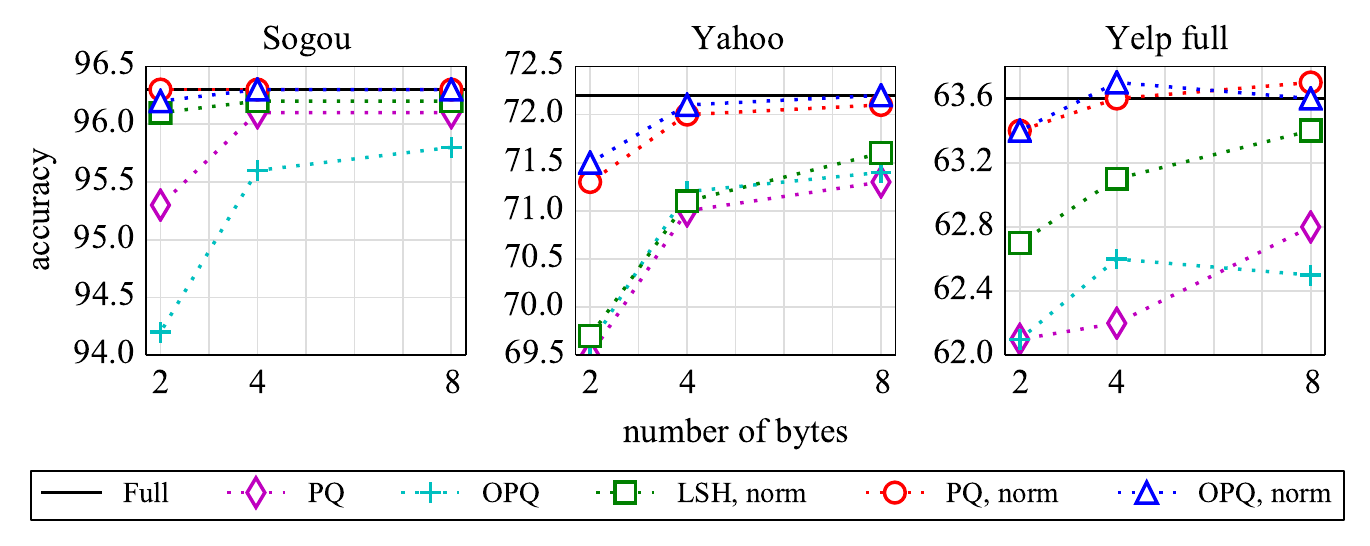} 
\caption{Accuracy as a function of the memory per vector/embedding on $3$ datasets from~\citet{Z15}. Note, an extra byte is required when we encode the norm explicitly ("norm"). 
}\label{fig:small}
\end{figure}

\paragraph{Compression techniques.}
We compare three popular methods used for similarity estimation with compact codes: LSH, PQ and OPQ on the
datasets released by \citet{Z15}.  Figure~\ref{fig:small} shows the accuracy as
a function of the number of bytes used per embedding, which corresponds to the number $k$ of subvectors 
in the case of PQ and OPQ. See more results in the appendix. As discussed in Section~\ref{sec:related}, LSH reproduces the cosine similarity and is therefore 
not adapted to un-normalized data. Therefore we only report results with normalization.  Once normalized, PQ and OPQ are
almost lossless even when using only $k=4$ subquantizers per embedding (equivalently, bytes). We observe in practice
that using $k=d/2$, \ie, half of the components of the embeddings, works well in
practice. In the rest of the paper and if not stated otherwise, we focus on this setting.  
The difference between the normalized versions of PQ and OPQ is limited and depends on the dataset. Therefore we
adopt the normalized PQ (NPQ) for the rest of this study, since it is faster to train.

\begin{figure}[t]
\centering
\includegraphics{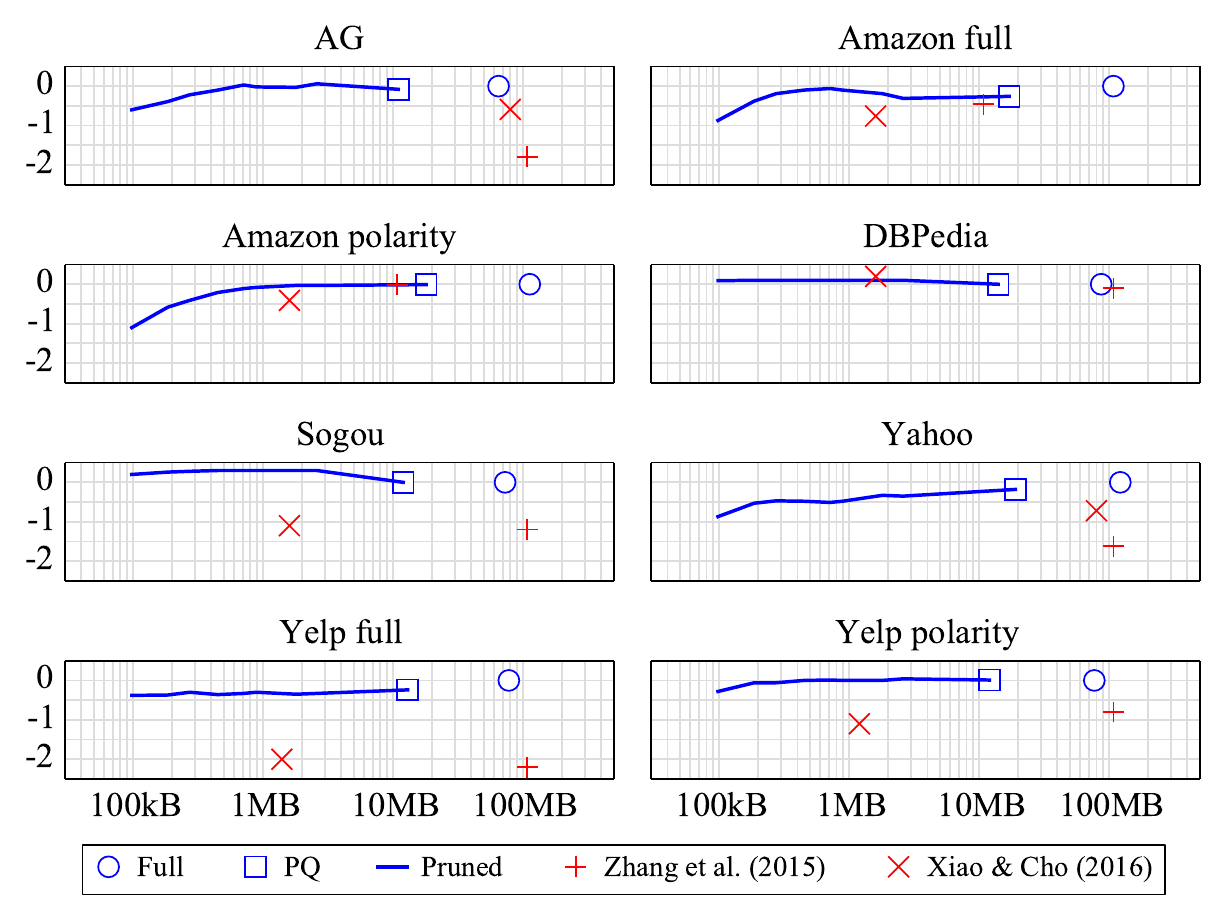} 
\vspace{-10pt}
\caption{Loss of accuracy as a function of the model size. We compare the compressed model
  with different level of pruning with NPQ and the full \texttt{fastText} model.
  We also compare with~\cite{Z15} and~\cite{xiao2016efficient}. Note that the size is
  in log scale.
\label{fig:pruning}
}
\end{figure}

\begin{table}[h!]
\centering
{\small
\begin{tabular}{lrr c lrr}
\toprule
word & Entropy & Norm &~~~& word & Entropy & Norm \\
\midrule
.   & 1 & 354 & & mediocre     & 1399& 1\\ 
,   & 2 & 176 & & disappointing& 454 & 2\\ 
the & 3 & 179 & & so-so        & 2809& 3\\ 
and & 4 & 1639& & lacks        & 1244& 4\\ 
i   & 5 & 2374& & worthless    & 1757& 5\\   
a   & 6 & 970 & & dreadful     & 4358& 6\\ 
to  & 7 & 1775& & drm          & 6395& 7\\ 
it  & 8 & 1956& & poorly       & 716 & 8\\ 
of  & 9 & 2815& & uninspired   & 4245& 9\\ 
this& 10& 3275& & worst        & 402 & 10\\
\bottomrule
\end{tabular}}
\caption{Best ranked words w.r.t. entropy (\emph{left}) and norm (\emph{right}) on the Amazon
  full review dataset. We give the rank for both criteria. 
  The norm ranking filters out words carrying little information. 
  }\label{tab:rank}
\end{table}
  
\paragraph{Pruning.}
Figure~\ref{fig:pruning} shows the performance of our model with different
sizes. We fix $k=d/2$ and use different pruning thresholds. NPQ offers a compression 
rate of $\times 10$ compared to the full model.  As the pruning becomes more agressive, 
the overall compression can increase up up to $\times 1,000$ with little
drop of performance and no additional overhead at test time. In fact, using a
smaller dictionary makes the model \emph{faster} at test time.  We
also compare with character-level Convolutional Neural Networks
(CNN)~\citep{Z15,xiao2016efficient}.  They are attractive models for text classification because they achieve
similar performance with less memory usage than linear
models~\citep{xiao2016efficient}.  Even though \texttt{fastText} with the
default setting uses more memory, NPQ is already on par with CNNs' memory
usage.  Note that CNNs are not quantized, and it would be worth seeing 
how much they can be quantized with no drop of performance. Such a study is beyond
the scope of this paper. 
Our pruning is based on the norm of the embeddings according to the guidelines of Section~\ref{sec:tricks}.
Table~\ref{tab:rank} compares the ranking obtained with norms to the
ranking obtained using entropy, which is commonly used in unsupervised
settings~\cite{stolcke2000entropy}. 

\begin{table}[t]
\centering
\begin{tabular}{l rl lll}
\toprule
Dataset & \multicolumn{2}{c}{full} & $64$KiB &  $32$KiB & $16$ KiB\\
\midrule
AG        & 65M    & 92.1 & 91.4 & 90.6 & 89.1 \\ 
Amazon full & 108M & 60.0 & 58.8 & 56.0 & 52.9 \\
Amazon pol. & 113M & 94.5 & 93.3 & 92.1 & 89.3\\
DBPedia   & 87M    & 98.4 & 98.2 & 98.1 & 97.4 \\
Sogou     & 73M    & 96.4 & 96.4 & 96.3 & 95.5\\
Yahoo     & 122M   & 72.1 & 70.0 & 69.0 & 69.2 \\
Yelp full & 78M    & 63.8 & 63.2 & 62.4 & 58.7 \\
Yelp pol. & 77M    & 95.7 & 95.3 & 94.9 & 93.2 \\
\midrule
Average diff.  [$\%$] & & 0 & -0.8 & -1.7 & -3.5 \\
\bottomrule
\end{tabular}
\caption{Performance on very small models. We use a quantization with $k=1$,
hashing and an extreme pruning. The last row shows the average drop of performance for different size.}
\label{tab:vsmall}
\end{table}

\paragraph{Extreme compression.}
Finally, in Table~\ref{tab:vsmall}, we explore the limit of quantized model
by looking at the performance obtained for models under $64$KiB. 
Surprisingly, even at $64$KiB and $32$KiB, the drop of performance is only around
$0.8\%$ and $1.7\%$ despite a compression rate of $\times 1,000-4,000$. 

\subsection{Large dataset: FlickrTag}

In this section, we explore the limit of compression algorithms on very large
datasets.  Similar to~\citet{J16}, we consider a hashtag prediction dataset
containing $312,116$ labels.  We set the minimum count for words at $10$,
leading to a dictionary of $1,427,667$ words. We take $10$M
buckets for n-grams and a hierarchical softmax. We refer to this dataset as FlickrTag. 

\paragraph{Output encoding.}
We are interested in
understanding how the performance degrades if  the classifier is also quantized
(\ie, the matrix $B$ in Eq.~\ref{eq:fasttext}) and when the pruning is at the
limit of the minimum number of features required to cover the full dataset. 

\begin{table}[h!]
\centering
\begin{tabular}{llcccc}
\toprule
Model & \ \ $k$ & norm & retrain & Acc. & Size \\
\midrule
full (uncompressed)         &     &   &   & 45.4 & 12\,GiB \\
\midrule
Input         & 128 &   &   & 45.0  & 1.7\,GiB\\
Input         & 128 & x &   & 45.3  & 1.8\,GiB\\
Input         & 128 & x & x & 45.5  & 1.8\,GiB\\
Input+Output  & 128 & x &   & 45.2  & 1.5\,GiB\\
Input+Output  & 128 & x & x & 45.4  & 1.5\,GiB\\
\bottomrule
\end{tabular}
\caption{FlickrTag: Influence of quantizing the output matrix on performance. We use PQ for quantization with an optional normalization.
We also retrain the output matrix after quantizing the input one. 
The "norm" refers to the separate encoding of the magnitude and angle, while "retrain" refers to the re-training bottom-up PQ method described in Section~\ref{sec:retrain}. 
}\label{tab:out}
\end{table}

Table~\ref{tab:out} shows that quantizing both the ``input'' matrix (\ie, $A$
    in Eq.~\ref{eq:fasttext}) and the ``output'' matrix  (\ie, $B$) does not degrade the
performance compared to the full model. We use embeddings with $d=256$
dimensions and use $k=d/2$ subquantizers. We do not use any text specific
tricks, which leads to a compression factor of $8$. Note that even if the
output matrix is not retrained over the embeddings, the performance is only
$0.2\%$ away from the full model.  As shown in the Appendix, using less
subquantizers significantly decreases the performance for a small memory gain.

\begin{table}[h!]
\centering
\begin{tabular}{lc cc cc cc}
\toprule
Model & full & \multicolumn{2}{c}{Entropy pruning} & \multicolumn{2}{c}{Norm pruning} & \multicolumn{2}{c}{Max-Cover pruning} \\
\cmidrule{3-4}\cmidrule(l){5-6}\cmidrule(l){7-8}
\#embeddings & 11.5M & 2M & 1M & 2M & 1M & 2M & 1M \\
Memory   & 12GiB & 297MiB & 174MiB & 305MiB & 179MiB & 305MiB & 179MiB \\
Coverage [$\%$] & 88.4 & 70.5 & 70.5 & 73.2 & 61.9 & 88.4  & 88.4\\
\midrule
Accuracy & 45.4 & 32.1 & 30.5 & 41.6 & 35.8 & 45.5 & 43.9 \\
\bottomrule
\end{tabular}
\caption{FlickrTag: Comparison of entropy pruning, norm pruning and max-cover pruning methods. We show the coverage of the
  test set for each method.}\label{tab:flickr_prune}
\end{table}

\paragraph{Pruning.}
Table~\ref{tab:flickr_prune} shows how the performance evolves with pruning.  We
measure this effect on top of a fully quantized model.  The full model misses
$11.6\%$ of the test set because of missing words (some documents are either
only composed of hashtags or have only rare words).  There are $312,116$
labels and thus it seems reasonable to keep embeddings in the order of the
million.  A naive pruning with $1$M features misses about $30-40\%$ of the test
set, leading to a significant drop of performance. On the other hand, even
though the max-coverage pruning approach was set on the train set, it does not
suffer from any coverage loss on the test set. This leads to a smaller drop of
performance. If the pruning is too aggressive, however, the coverage
decreases significantly.

\section{Future Work}

It may be possible to obtain further reduction of the model size in the future.
One idea is to condition the size of the vectors (both for the input features
 and the labels) based on their
frequency~\citep{chen2015strategies,grave2016efficient}. For example, it is
probably not worth representing the rare labels by full 256-dimensional vectors
in the case of the FlickrTag dataset. Thus, conditioning the vector size on the
frequency and norm seems like an interesting direction to explore
in the future.

We may also consider combining the entropy and norm pruning criteria:
instead of keeping the features in the model based just on the frequency or
the norm, we can use both to keep a good set of features. This could
help to keep features that are both frequent and discriminative, and thereby to
reduce the coverage problem that we have observed.

Additionally, instead of pruning out the less useful features, we can decompose
them into smaller units~\citep{mikolov2012subword}. For example, this can be achieved
by splitting every non-discriminative word into a sequence of
character trigrams. This could help in cases where training and test examples
are very short (for example just a single word).

\section{Conclusion}

In this paper, we have presented several simple techniques to reduce, by several orders of magnitude, the memory complexity
of certain text classifiers without sacrificing accuracy nor speed.
This is achieved by applying discriminative pruning which aims to keep only important features in the trained model,
and by performing quantization of the weight matrices and hashing of the dictionary.

We will publish the code as an extension of the \texttt{fastText} library.
We hope that our work will serve as a baseline to the research community, where
there is an increasing interest for comparing the performance of various deep
learning text classifiers for a given number of parameters. 
Overall, compared to recent work based on convolutional neural networks,
  \texttt{fastText.zip} is often more accurate, while requiring several orders
  of magnitude less
time to train on common CPUs, and incurring a fraction of the memory
complexity.

\bibliography{egbib}

\begin{thebibliography}{43}
\providecommand{\natexlab}[1]{#1}
\providecommand{\url}[1]{\texttt{#1}}
\expandafter\ifx\csname urlstyle\endcsname\relax
  \providecommand{\doi}[1]{doi: #1}\else
  \providecommand{\doi}{doi: \begingroup \urlstyle{rm}\Url}\fi

\bibitem[Agarwal et~al.(2014)Agarwal, Chapelle, Dud{\'\i}k, and Langford]{A14}
Alekh Agarwal, Olivier Chapelle, Miroslav Dud{\'\i}k, and John Langford.
\newblock A reliable effective terascale linear learning system.
\newblock \emph{Journal of Machine Learning Research}, 15\penalty0
  (1):\penalty0 1111--1133, 2014.

\bibitem[Bach et~al.(2012)Bach, Jenatton, Mairal, and Obozinski]{BJ12}
Francis Bach, Rodolphe Jenatton, Julien Mairal, and Guillaume Obozinski.
\newblock Optimization with sparsity-inducing penalties.
\newblock \emph{Foundations and Trends{\textregistered} in Machine Learning},
  4\penalty0 (1):\penalty0 1--106, 2012.

\bibitem[Badanidiyuru et~al.(2014)Badanidiyuru, Mirzasoleiman, Karbasi, and
  Krause]{badanidiyuru2014streaming}
Ashwinkumar Badanidiyuru, Baharan Mirzasoleiman, Amin Karbasi, and Andreas
  Krause.
\newblock Streaming submodular maximization: Massive data summarization on the
  fly.
\newblock In \emph{SIGKDD}, pp.\  671--680. ACM, 2014.

\bibitem[Bateni et~al.(2010)Bateni, Hajiaghayi, and
  Zadimoghaddam]{bateni2010submodular}
Mohammad~Hossein Bateni, Mohammad~Taghi Hajiaghayi, and Morteza Zadimoghaddam.
\newblock Submodular secretary problem and extensions.
\newblock In \emph{Approximation, Randomization, and Combinatorial
  Optimization. Algorithms and Techniques}, pp.\  39--52. Springer, 2010.

\bibitem[Charikar(2002)]{C02}
Moses~S. Charikar.
\newblock Similarity estimation techniques from rounding algorithms.
\newblock In \emph{STOC}, pp.\  380--388, May 2002.

\bibitem[Chen et~al.(2015)Chen, Grangier, and Auli]{chen2015strategies}
Welin Chen, David Grangier, and Michael Auli.
\newblock Strategies for training large vocabulary neural language models.
\newblock \emph{arXiv preprint arXiv:1512.04906}, 2015.

\bibitem[Chierichetti et~al.(2010)Chierichetti, Kumar, and
  Tomkins]{chierichetti2010max}
Flavio Chierichetti, Ravi Kumar, and Andrew Tomkins.
\newblock Max-cover in map-reduce.
\newblock In \emph{International Conference on World Wide Web}, 2010.

\bibitem[Courbariaux et~al.(2016)Courbariaux, Hubara, Soudry, El-Yaniv, and
  Bengio]{CHSEB16}
Matthieu Courbariaux, Itay Hubara, Daniel Soudry, Ran El-Yaniv, and Yoshua
  Bengio.
\newblock Binarized neural networks: Training neural networks with weights and
  activations constrained to +1 or -1.
\newblock \emph{arXiv preprint arXiv:1602.02830}, 2016.

\bibitem[Datar et~al.(2004)Datar, Immorlica, Indyk, and Mirrokni]{DIIM04}
M.~Datar, N.~Immorlica, P.~Indyk, and V.S. Mirrokni.
\newblock Locality-sensitive hashing scheme based on p-stable distributions.
\newblock In \emph{Proceedings of the Symposium on Computational Geometry},
  pp.\  253--262, 2004.

\bibitem[Deerwester et~al.(1990)Deerwester, Dumais, Furnas, Landauer, and
  Harshman]{deerwester1990indexing}
Scott Deerwester, Susan~T Dumais, George~W Furnas, Thomas~K Landauer, and
  Richard Harshman.
\newblock Indexing by latent semantic analysis.
\newblock \emph{Journal of the American society for information science}, 1990.

\bibitem[Denil et~al.(2013)Denil, Shakibi, Dinh, Ranzato, and
  de~Freitas]{DSDRF13}
Misha Denil, Babak Shakibi, Laurent Dinh, Marc-Aurelio Ranzato, and Nando
  et~all de~Freitas.
\newblock Predicting parameters in deep learning.
\newblock In \emph{NIPS}, pp.\  2148--2156, 2013.

\bibitem[Feige(1998)]{feige1998threshold}
Uriel Feige.
\newblock A threshold of ln n for approximating set cover.
\newblock \emph{JACM}, 45\penalty0 (4):\penalty0 634--652, 1998.

\bibitem[Ge et~al.(2013)Ge, He, Ke, and Sun]{GHKS13}
Tiezheng Ge, Kaiming He, Qifa Ke, and Jian Sun.
\newblock Optimized product quantization for approximate nearest neighbor
  search.
\newblock In \emph{CVPR}, June 2013.

\bibitem[Gong \& Lazebnik(2011)Gong and Lazebnik]{GL11}
Yunchao Gong and Svetlana Lazebnik.
\newblock Iterative quantization: A procrustean approach to learning binary
  codes.
\newblock In \emph{CVPR}, June 2011.

\bibitem[Gong et~al.(2014)Gong, Liu, Yang, and Bourdev]{GLYB14}
Yunchao Gong, Liu Liu, Ming Yang, and Lubomir Bourdev.
\newblock Compressing deep convolutional networks using vector quantization.
\newblock \emph{arXiv preprint arXiv:1412.6115}, 2014.

\bibitem[Grave et~al.(2016)Grave, Joulin, Ciss{\'e}, Grangier, and
  J{\'e}gou]{grave2016efficient}
Edouard Grave, Armand Joulin, Moustapha Ciss{\'e}, David Grangier, and
  Herv{\'e} J{\'e}gou.
\newblock Efficient softmax approximation for gpus.
\newblock \emph{arXiv preprint arXiv:1609.04309}, 2016.

\bibitem[Han et~al.(2016)Han, Mao, and Dally]{HMD16}
Song Han, Huizi Mao, and William~J Dally.
\newblock Deep compression: Compressing deep neural networks with pruning,
  trained quantization and huffman coding.
\newblock In \emph{ICLR}, 2016.

\bibitem[J\'egou et~al.(2008)J\'egou, Douze, and Schmid]{JDS08}
Herv\'e J\'egou, Matthijs Douze, and Cordelia Schmid.
\newblock Hamming embedding and weak geometric consistency for large scale
  image search.
\newblock In \emph{ECCV}, October 2008.

\bibitem[Jegou et~al.(2011)Jegou, Douze, and Schmid]{JDS11}
Herv\'{e} Jegou, Matthijs Douze, and Cordelia Schmid.
\newblock Product quantization for nearest neighbor search.
\newblock \emph{IEEE Trans. PAMI}, January 2011.

\bibitem[Joachims(1998)]{joachims1998text}
Thorsten Joachims.
\newblock \emph{Text categorization with support vector machines: Learning with
  many relevant features}.
\newblock Springer, 1998.

\bibitem[Joulin et~al.(2016)Joulin, Grave, Bojanowski, and Mikolov]{J16}
Armand Joulin, Edouard Grave, Piotr Bojanowski, and Tomas Mikolov.
\newblock Bag of tricks for efficient text classification.
\newblock \emph{arXiv preprint arXiv:1607.01759}, 2016.

\bibitem[LeCun et~al.(1990)LeCun, Denker, and Solla]{LDS90}
Yann LeCun, John~S Denker, and Sara~A Solla.
\newblock Optimal brain damage.
\newblock \emph{NIPS}, 2:\penalty0 598--605, 1990.

\bibitem[Lin et~al.(2015)Lin, Courbariaux, Memisevic, and Bengio]{LCMB15}
Zhouhan Lin, Matthieu Courbariaux, Roland Memisevic, and Yoshua Bengio.
\newblock Neural networks with few multiplications.
\newblock \emph{arXiv preprint arXiv:1510.03009}, 2015.

\bibitem[McCallum \& Nigam(1998)McCallum and Nigam]{mccallum1998comparison}
Andrew McCallum and Kamal Nigam.
\newblock A comparison of event models for naive bayes text classification.
\newblock In \emph{AAAI workshop on learning for text categorization}, 1998.

\bibitem[Meier et~al.(2008)Meier, Van De~Geer, and B{\"u}hlmann]{M08}
Lukas Meier, Sara Van De~Geer, and Peter B{\"u}hlmann.
\newblock The group lasso for logistic regression.
\newblock \emph{Journal of the Royal Statistical Society: Series B (Statistical
  Methodology)}, 70\penalty0 (1):\penalty0 53--71, 2008.

\bibitem[Mikolov(2012)]{mikolov2012}
Tomas Mikolov.
\newblock Statistical language models based on neural networks.
\newblock In \emph{PhD thesis}. VUT Brno, 2012.

\bibitem[Mikolov et~al.(2012)Mikolov, Sutskever, Deoras, Le, Kombrink, and
  Cernocky]{mikolov2012subword}
Tomas Mikolov, Ilya Sutskever, Anoop Deoras, Hai-Son Le, Stefan Kombrink, and
  J~Cernocky.
\newblock Subword language modeling with neural networks.
\newblock \emph{preprint}, 2012.

\bibitem[Neyshabur \& Srebro(2015)Neyshabur and Srebro]{NS15}
Behnam Neyshabur and Nathan Srebro.
\newblock On symmetric and asymmetric lshs for inner product search.
\newblock In \emph{ICML}, pp.\  1926--1934, 2015.

\bibitem[Norouzi \& Fleet(2013)Norouzi and Fleet]{NF13}
Mohammad Norouzi and David Fleet.
\newblock Cartesian k-means.
\newblock In \emph{CVPR}, June 2013.

\bibitem[Pang \& Lee(2008)Pang and Lee]{pang2008opinion}
Bo~Pang and Lillian Lee.
\newblock Opinion mining and sentiment analysis.
\newblock \emph{Foundations and trends in information retrieval}, 2008.

\bibitem[Sablayrolles et~al.(2016)Sablayrolles, Douze, J{\'e}gou, and
  Usunier]{S16}
Alexandre Sablayrolles, Matthijs Douze, Herv{\'e} J{\'e}gou, and Nicolas
  Usunier.
\newblock How should we evaluate supervised hashing?
\newblock \emph{arXiv preprint arXiv:1609.06753}, 2016.

\bibitem[S{\'a}nchez \& Perronnin(2011)S{\'a}nchez and Perronnin]{SP11}
Jorge S{\'a}nchez and Florent Perronnin.
\newblock High-dimensional signature compression for large-scale image
  classification.
\newblock In \emph{CVPR}, 2011.

\bibitem[Shrivastava \& Li(2014)Shrivastava and Li]{SL14}
Anshumali Shrivastava and Ping Li.
\newblock Asymmetric {LSH} for sublinear time maximum inner product search.
\newblock In \emph{NIPS}, pp.\  2321--2329, 2014.

\bibitem[Stolcke(2000)]{stolcke2000entropy}
Andreas Stolcke.
\newblock Entropy-based pruning of backoff language models.
\newblock \emph{arXiv preprint cs/0006025}, 2000.

\bibitem[Talbot \& Brants(2008)Talbot and Brants]{talbot2008randomized}
David Talbot and Thorsten Brants.
\newblock Randomized language models via perfect hash functions.
\newblock In \emph{ACL}, 2008.

\bibitem[Thomee et~al.(2016)Thomee, Shamma, Friedland, Elizalde, Ni, Poland,
  Borth, and Li]{T15}
Bart Thomee, David~A Shamma, Gerald Friedland, Benjamin Elizalde, Karl Ni,
  Douglas Poland, Damian Borth, and Li-Jia Li.
\newblock Yfcc100m: The new data in multimedia research.
\newblock In \emph{Communications of the ACM}, 2016.

\bibitem[Wang et~al.(2014)Wang, Shen, Song, and Ji]{WSSJ14}
Jingdong Wang, Heng~Tao Shen, Jingkuan Song, and Jianqiu Ji.
\newblock Hashing for similarity search: A survey.
\newblock \emph{arXiv preprint arXiv:1408.2927}, 2014.

\bibitem[Wang et~al.(2015)Wang, Liu, Kumar, and Chang]{WLKC15}
Jun Wang, Wei Liu, Sanjiv Kumar, and Shih{-}Fu Chang.
\newblock Learning to hash for indexing big data - {A} survey.
\newblock \emph{CoRR}, abs/1509.05472, 2015.

\bibitem[Wang \& Manning(2012)Wang and Manning]{W12}
Sida Wang and Christopher~D Manning.
\newblock Baselines and bigrams: Simple, good sentiment and topic
  classification.
\newblock In \emph{ACL}, 2012.

\bibitem[Weinberger et~al.(2009)Weinberger, Dasgupta, Langford, Smola, and
  Attenberg]{W09}
Kilian~Q Weinberger, Anirban Dasgupta, John Langford, Alex Smola, and Josh
  Attenberg.
\newblock Feature hashing for large scale multitask learning.
\newblock In \emph{ICML}, 2009.

\bibitem[Weiss et~al.(2009)Weiss, Torralba, and Fergus]{WTF09}
Yair Weiss, Antonio Torralba, and Rob Fergus.
\newblock Spectral hashing.
\newblock In \emph{NIPS}, December 2009.

\bibitem[Xiao \& Cho(2016)Xiao and Cho]{xiao2016efficient}
Yijun Xiao and Kyunghyun Cho.
\newblock Efficient character-level document classification by combining
  convolution and recurrent layers.
\newblock \emph{arXiv preprint arXiv:1602.00367}, 2016.

\bibitem[Zhang et~al.(2015)Zhang, Zhao, and LeCun]{Z15}
Xiang Zhang, Junbo Zhao, and Yann LeCun.
\newblock Character-level convolutional networks for text classification.
\newblock In \emph{NIPS}, 2015.

\end{thebibliography}
\bibliographystyle{iclr2017_conference}

\newpage
\section*{Appendix}

In the appendix, we show some additional results. The model used in these
experiments only had $1$M ngram buckets. In Table~\ref{tab:prem}, we show a
thorough comparison of LSH, PQ and OPQ on $8$ different datasets.
Table~\ref{tab:cnn} summarizes the comparison with CNNs in terms of accuracy
and size.  Table~\ref{tab:bloom} show a thorough comparison of the hashing
trick and the Bloom filters.

\begin{table}[h!]
{\small
\hspace{-60pt}\begin{tabular}{llc c@{\mys}c c@{\mys}c c@{\mys}c c@{\mys}c c@{\mys}c c@{\mys}c c@{\mys}c c@{\mys}c}
\toprule
Quant. & k & \rotatebox{75}{norm} & \multicolumn{2}{c}{AG} & \multicolumn{2}{c}{Amz. f.} & \multicolumn{2}{c}{Amz. p.} & \multicolumn{2}{c}{DBP} & \multicolumn{2}{c}{Sogou} & \multicolumn{2}{c}{Yah.} & \multicolumn{2}{c}{Yelp f.} & \multicolumn{2}{c}{Yelp p.} 
\\
\midrule
\multicolumn{3}{l}{full}        & 92.1 & 36M & 59.8 & 97M & 94.5 & 104M & 98.4 & 67M & 96.3 & 47M & 72 & 120M & 63.7 & 56M & 95.7 & 53M \\
\multicolumn{3}{l}{full,nodict} & 92.1 & 34M & 59.9 & 78M & 94.5 & 83M & 98.4 & 56M & 96.3 & 42M & 72.2 & 91M & 63.6 & 48M & 95.6 & 46M \\
\midrule
LSH         & 8 &   & 88.7 & 8.5M & 51.3 & 20M & 90.3 & 21M & 92.7 & 14M & 94.2 & 11M & 54.8 & 23M & 56.7 & 12M & 92.2 & 12M \\
PQ          & 8 &   & 91.7 & 8.5M & 59.3 & 20M & 94.4 & 21M & 97.4 & 14M & 96.1 & 11M & 71.3 & 23M & 62.8 & 12M & 95.4 & 12M \\
OPQ         & 8 &   & 91.9 & 8.5M & 59.3 & 20M & 94.4 & 21M & 96.9 & 14M & 95.8 & 11M & 71.4 & 23M & 62.5 & 12M & 95.4 & 12M \\
LSH         & 8 & x & 91.9 & 9.5M & 59.4 & 22M & 94.5 & 24M & 97.8 & 16M & 96.2 & 12M & 71.6 & 26M & 63.4 & 14M & 95.6 & 13M \\
PQ          & 8 & x & 92.0 & 9.5M & 59.8 & 22M & 94.5 & 24M & 98.4 & 16M & 96.3 & 12M & 72.1 & 26M & 63.7 & 14M & 95.6 & 13M \\
OPQ         & 8 & x & 92.1 & 9.5M & 59.9 & 22M & 94.5 & 24M & 98.4 & 16M & 96.3 & 12M & 72.2 & 26M & 63.6 & 14M & 95.6 & 13M \\
\midrule
LSH         & 4 &   & 88.3 & 4.3M & 50.5 & 9.7M & 88.9 & 11M & 91.6 & 7.0M & 94.3 & 5.3M & 54.6 & 12M & 56.5 & 6.0M & 92.9 & 5.7M \\
PQ          & 4 &   & 91.6 & 4.3M & 59.2 & 9.7M & 94.4 & 11M & 96.3 & 7.0M & 96.1 & 5.3M & 71.0 & 12M & 62.2 & 6.0M & 95.4 & 5.7M \\
OPQ         & 4 &   & 91.7 & 4.3M & 59.0 & 9.7M & 94.4 & 11M & 96.9 & 7.0M & 95.6 & 5.3M & 71.2 & 12M & 62.6 & 6.0M & 95.4 & 5.7M \\
LSH         & 4 & x & 92.1 & 5.3M & 59.2 & 13M & 94.4 & 13M & 97.7 & 8.8M & 96.2 & 6.6M & 71.1 & 15M & 63.1 & 7.4M & 95.5 & 7.2M \\
PQ          & 4 & x & 92.1 & 5.3M & 59.8 & 13M & 94.5 & 13M & 98.4 & 8.8M & 96.3 & 6.6M & 72.0 & 15M & 63.6 & 7.5M & 95.6 & 7.2M \\
OPQ         & 4 & x & 92.2 & 5.3M & 59.8 & 13M & 94.5 & 13M & 98.3 & 8.8M & 96.3 & 6.6M & 72.1 & 15M & 63.7 & 7.5M & 95.6 & 7.2M \\
\midrule 
LSH         & 2 &   & 87.7 & 2.2M & 50.1 & 4.9M & 88.9 & 5.2M & 90.6 & 3.5M & 93.9 & 2.7M & 51.4 & 5.7M & 56.6 & 3.0M & 91.3 & 2.9M \\
PQ          & 2 &   & 91.1 & 2.2M & 58.7 & 4.9M & 94.4 & 5.2M & 87.1 & 3.6M & 95.3 & 2.7M & 69.5 & 5.7M & 62.1 & 3.0M & 95.4 & 2.9M \\
OPQ         & 2 &   & 91.4 & 2.2M & 58.2 & 4.9M & 94.3 & 5.2M & 91.6 & 3.6M & 94.2 & 2.7M & 69.6 & 5.7M & 62.1 & 3.0M & 95.4 & 2.9M \\
LSH         & 2 & x & 91.8 & 3.2M & 58.6 & 7.3M & 94.3 & 7.8M & 97.1 & 5.3M & 96.1 & 4.0M & 69.7 & 8.6M & 62.7 & 4.5M & 95.5 & 4.3M \\
PQ          & 2 & x & 91.9 & 3.2M & 59.6 & 7.3M & 94.5 & 7.8M & 98.1 & 5.3M & 96.3 & 4.0M & 71.3 & 8.6M & 63.4 & 4.5M & 95.6 & 4.3M \\
OPQ         & 2 & x & 92.1 & 3.2M & 59.5 & 7.3M & 94.5 & 7.8M & 98.1 & 5.3M & 96.2 & 4.0M & 71.5 & 8.6M & 63.4 & 4.5M & 95.6 & 4.3M \\
\bottomrule
\end{tabular}
}
\caption{Comparison between standard quantization methods. The original model has a dimensionality of $8$ and $2$M buckets.
Note that all of the methods are without dictionary.}\label{tab:prem}
\end{table}

\begin{table}[h!]
{\small
\hspace{-50pt}\begin{tabular}{lr c@{\mys}c c@{\mys}c c@{\mys}c c@{\mys}c c@{\mys}c c@{\mys}c c@{\mys}c c@{\mys}c}
\toprule
 k & co & \multicolumn{2}{c}{AG} & \multicolumn{2}{c}{Amz. f.} & \multicolumn{2}{c}{Amz. p.} & \multicolumn{2}{c}{DBP} & \multicolumn{2}{c}{Sogou} & \multicolumn{2}{c}{Yah.} & \multicolumn{2}{c}{Yelp f.} & \multicolumn{2}{c}{Yelp p.} 
\\ \midrule
\multicolumn{2}{l}{full, nodict} & 92.1 & 34M & 59.8 & 78M & 94.5 & 83M & 98.4 & 56M & 96.3 & 42M & 72.2 & 91M & 63.7 & 48M & 95.6 & 46M \\
8  & full & 92.0 & 9.5M & 59.8 & 22M & 94.5 & 24M & 98.4 & 16M & 96.3 & 12M & 72.1 & 26M & 63.7 & 14M & 95.6 & 13M \\
4 & full & 92.1 & 5.3M & 59.8 & 13M & 94.5 & 13M & 98.4 & 8.8M & 96.3 & 6.6M & 72 & 15M & 63.6 & 7.5M & 95.6 & 7.2M \\
2 & full & 91.9 & 3.2M & 59.6 & 7.3M & 94.5 & 7.8M & 98.1 & 5.3M & 96.3 & 4.0M & 71.3 & 8.6M & 63.4 & 4.5M & 95.6 & 4.3M \\
\midrule
8&200K & 92.0 & 2.5M & 59.7 & 2.5M & 94.3 & 2.5M & 98.5 & 2.5M & 96.6 & 2.5M & 71.8 & 2.5M & 63.3 & 2.5M & 95.6 & 2.5M \\
8&100K & 91.9 & 1.3M & 59.5 & 1.3M & 94.3 & 1.3M & 98.5 & 1.3M & 96.6 & 1.3M & 71.6 & 1.3M & 63.4 & 1.3M & 95.6 & 1.3M \\
8&50K & 91.7 & 645K & 59.7 & 645K & 94.3 & 644K & 98.5 & 645K & 96.6 & 645K & 71.5 & 645K & 63.2 & 645K & 95.6 & 644K \\
8&10K & 91.3 & 137K & 58.6 & 137K & 93.2 & 137K & 98.5 & 137K & 96.5 & 137K & 71.3 & 137K & 63.3 & 137K & 95.4 & 137K \\
\midrule
4&200K & 92.0 & 1.8M & 59.7 & 1.8M & 94.3 & 1.8M & 98.5 & 1.8M & 96.6 & 1.8M & 71.7 & 1.8M & 63.3 & 1.8M & 95.6 & 1.8M \\
4&100K & 91.9 & 889K & 59.5 & 889K & 94.4 & 889K & 98.5 & 889K & 96.6 & 889K & 71.7 & 889K & 63.4 & 889K & 95.6 & 889K \\
4&50K & 91.7 & 449K & 59.6 & 449K & 94.3 & 449K & 98.5 & 450K & 96.6 & 449K & 71.4 & 450K & 63.2 & 449K & 95.5 & 449K \\
4&10K & 91.5 & 98K & 58.6 & 98K & 93.2 & 98K & 98.5 & 98K & 96.5 & 98K & 71.2 & 98K & 63.3 & 98K & 95.4 & 98K \\
\midrule
2&200K & 91.9 & 1.4M & 59.6 & 1.4M & 94.3 & 1.4M & 98.4 & 1.4M & 96.5 & 1.4M & 71.5 & 1.4M & 63.2 & 1.4M & 95.5 & 1.4M \\
2&100K & 91.6 & 693K & 59.5 & 693K & 94.3 & 693K & 98.4 & 694K & 96.6 & 693K & 71.1 & 694K & 63.2 & 693K & 95.6 & 693K \\
2&50K & 91.6 & 352K & 59.6 & 352K & 94.3 & 352K & 98.4 & 352K & 96.5 & 352K & 71.1 & 352K & 63.2 & 352K & 95.6 & 352K \\
2&10K & 91.3 & 78K & 58.5 & 78K & 93.2 & 78K & 98.4 & 79K & 96.5 & 78K & 70.8 & 78K & 63.2 & 78K & 95.3 & 78K \\
\bottomrule
\end{tabular}
}
\caption{Comparison with different quantization and level of pruning. ``co'' is the cut-off parameter of the pruning.}
\end{table}

\begin{table}[h!]
\centering {\small
\begin{tabular}{l cc cc cc}
\toprule
Dataset & \multicolumn{2}{c}{\cite{Z15}} & \multicolumn{2}{c}{\cite{xiao2016efficient}} & 
 \multicolumn{2}{c}{\texttt{fastText}+PQ, $k=d/2$}
\\ \midrule
AG      & 90.2 & 108M  & 91.4 & 80M  & 91.9 & 889K \\ 
Amz. f. & 59.5 & 10.8M & 59.2 & 1.6M & 59.6 & 449K \\
Amz. p. & 94.5 & 10.8M & 94.1 & 1.6M & 94.3 & 449K \\
DBP     & 98.3 & 108M  & 98.6 & 1.2M & 98.5 & 98K \\
Sogou   & 95.1 & 108M  & 95.2 & 1.6M & 96.5 & 98K \\
Yah.    & 70.5 & 108M  & 71.4 & 80M  & 71.7 & 889K \\
Yelp f. & 61.6 & 108M  & 61.8 & 1.4M & 63.3 & 98K \\
Yelp p. & 94.8 & 108M  & 94.5 & 1.2M & 95.5 & 449K \\
\bottomrule
\end{tabular}
}
\caption{Comparison between CNNs and \texttt{fastText} with and without 
quantization. The numbers for~\cite{Z15} 
are reported from~\citet{xiao2016efficient}. Note that for the CNNs, we report
the size of the model under the assumption that they use float32 storage. For \texttt{fastText}(+PQ) we report the memory used in RAM at test time.}
\label{tab:cnn}
\end{table}

\begin{table}[h!]
\hspace{-55pt}
{\small
\begin{tabular}{lcr c@{\mys}r@{\mysb} c@{\mys}r@{\mysb} c@{\mys}r@{\mysb} c@{\mys}r@{\mysb} c@{\mys}r@{\mysb} c@{\mys}r@{\mysb} c@{\mys}r@{\mysb} c@{\mys}r}
\toprule
Quant. & \rotatebox{75}{Bloom} & co\ \  & \multicolumn{2}{c}{AG} & \multicolumn{2}{c}{Amz. f.} & \multicolumn{2}{c}{Amz. p.} & 
\multicolumn{2}{c}{DBP} &\multicolumn{2}{c}{Sogou} & \multicolumn{2}{c}{Yah.} & \multicolumn{2}{c}{Yelp f.} & \multicolumn{2}{c}{Yelp p.} 
\\ \midrule
\multicolumn{2}{l}{full,nodict}  && 92.1 & 34M & 59.8 & 78M & 94.5 & 83M & 98.4 & 56M & 96.3 & 42M & 72.2 & 91M & 63.7 & 48M & 95.6 & 46M \\
\midrule
NPQ         & &200K & 91.9 & 1.4M & 59.6 & 1.4M & 94.3 & 1.4M & 98.4 & 1.4M & 96.5 & 1.4M & 71.5 & 1.4M & 63.2 & 1.4M & 95.5 & 1.4M \\
NPQ         &x&200K & 92.2 & 830K & 59.3 & 830K & 94.1 & 830K & 98.4 & 830K & 96.5 & 830K & 70.7 & 830K & 63.0 & 830K & 95.5 & 830K \\
NPQ         & &100K & 91.6 & 693K & 59.5 & 693K & 94.3 & 693K & 98.4 & 694K & 96.6 & 693K & 71.1 & 694K & 63.2 & 693K & 95.6 & 693K \\
NPQ         &x&100K & 91.8 & 420K & 59.1 & 420K & 93.9 & 420K & 98.4 & 420K & 96.5 & 420K & 70.6 & 420K & 62.8 & 420K & 95.3 & 420K \\
NPQ         & &50K & 91.6 & 352K & 59.6 & 352K & 94.3 & 352K & 98.4 & 352K & 96.5 & 352K & 71.1 & 352K & 63.2 & 352K & 95.6 & 352K \\
NPQ         &x&50K & 91.5 & 215K & 58.8 & 215K & 93.6 & 215K & 98.3 & 215K & 96.5 & 215K & 70.1 & 215K & 62.7 & 215K & 95.1 & 215K \\
NPQ         & &10K & 91.3 & 78K & 58.5 & 78K & 93.2 & 78K & 98.4 & 79K & 96.5 & 78K & 70.8 & 78K & 63.2 & 78K & 95.3 & 78K \\
NPQ         &x&10K & 90.8 & 51K & 56.8 & 51K & 91.7 & 51K & 98.1 & 51K & 96.1 & 51K & 68.7 & 51K & 61.7 & 51K & 94.5 & 51K \\
\bottomrule
\end{tabular}
}
\vspace{-4pt}
\caption{Comparison with and without Bloom filters. For NPQ, we set $d=8$ and $k=2$.}\label{tab:bloom}
\end{table}

\begin{table}[h!]
\centering {\small
\begin{tabular}{lrcccc}
\toprule
Model                               & k   & norm  & retrain  & Acc. & Size \\
\midrule
full                                &     &    &   & 45.4 & 12G \\
\midrule
Input                               & 128 &    &   & 45.0 & 1.7G\\
Input                               & 128 & x  &   & 45.3 & 1.8G\\
Input                               & 128 & x  & x & 45.5 & 1.8G\\
Input+Output                        & 128 & x  &   & 45.2 & 1.5G\\
Input+Output                        & 128 & x  & x & 45.4 & 1.5G\\
Input+Output, co=2M                 & 128 & x  & x & 45.5 & 305M\\
Input+Output, n co=1M               & 128 & x  & x & 43.9 & 179M\\
\midrule  
Input                               &  64 &    &   & 44.0 & 1.1G\\
Input                               &  64 & x  &   & 44.7 & 1.1G\\
Input                               &  64 & x  &   & 44.9 & 1.1G\\
Input+Output                        &  64 & x  &   & 44.6 & 784M\\
Input+Output                        &  64 & x  & x & 44.8 & 784M\\
Input+Output, co=2M                 &  64 & x  &   & 42.5 & 183M\\
Input+Output, co=1M                 &  64 & x  &   & 39.9 & 118M\\
Input+Output, co=2M                 &  64 & x  & x & 45.0 & 183M\\
Input+Output, co=1M                 &  64 & x  & x & 43.4 & 118M\\
\midrule 
Input                               &  32 &    &   & 40.5 & 690M\\
Input                               &  32 & x  &   & 42.4 & 701M\\
Input                               &  32 & x  & x & 42.9 & 701M\\
Input+Output                        &  32 & x  &   & 42.3 & 435M\\
Input+Output                        &  32 & x  & x & 42.8 & 435M\\
Input+Output, co=2M                 &  32 & x  &   & 35.0 & 122M\\
Input+Output, co=1M                 &  32 & x  &   & 32.6 & 88M\\
Input+Output, co=2M                 &  32 & x  & x & 43.3 & 122M\\
Input+Output, co=1M                 &  32 & x  & x & 41.6 & 88M\\
\bottomrule
\end{tabular}
}
\caption{FlickrTag: Comparison for a large dataset of (i) different quantization methods and parameters, (ii) with or without re-training.}\label{tab:flickr}
\end{table}

\end{document}